\documentclass[10pt,twocolumn,letterpaper]{article}

\usepackage[pagenumbers]{cvpr} 

\usepackage{booktabs}
\usepackage{multirow}
\usepackage{array}
\usepackage{lipsum}
\usepackage{makecell}
\usepackage{mathtools}
\usepackage[table]{xcolor}
\usepackage{titletoc}
\usepackage{minitoc}
\definecolor{cvprblue}{rgb}{0.21,0.49,0.74}
\usepackage[pagebackref,breaklinks,colorlinks,allcolors=cvprblue]{hyperref}

\def\paperID{*****} 
\def\confName{CVPR}
\def\confYear{2026}

\title{DTG-Restore: Training-Free Diffusion Refinement for Generative Video Super-Resolution}

\author{
Hidir Yesiltepe$^{1}$ \quad Koutilya PNVR$^{2}$ \quad Gaurav Pathak$^{2}$ \quad Navaneeth Bodla$^{2}$ \\
Bharat Singh$^{2}$ \quad Pinar Yanardag$^{\dagger 1}$ \quad Jinrong Xie$^{\dagger 2}$ \\
$^{1}$Virginia Tech \quad $^{2}$Adobe \\
}

\begin{document}
\maketitle
\begingroup
\renewcommand\thefootnote{}
\footnotetext{${}^{\dagger}$Equal supervision}
\endgroup
\begin{abstract}
Recent progress in video diffusion models has enabled remarkable generative fidelity, yet leveraging these priors for restoration remains limited by the strong coupling between conditional and unconditional branches in standard classifier-free guidance. We introduce a \emph{training-free} framework that enhances distorted and low-resolution videos by decoupling these signals in time. Our proposed \textbf{Decoupled Time Guidance (DTG)} evaluates the unconditional branch at a cleaner diffusion timestep, providing a lookahead prior that preserves geometry while suppressing replication of warped content. This temporal bias is annealed throughout sampling, allowing the model to transition from structure correction to detail refinement without retraining. Combined with any off-the-shelf restoration module in a plug-and-play manner, our approach improves perceptual coherence and restores plausible structure in AI-generated and real-world videos alike. To facilitate evaluation, we curate \textbf{GenWarp480}, a benchmark of 4{,}400 distorted 480p videos synthesized from diverse text-to-video models. GenWarp480 focuses on characteristic generative degradations such as warped faces, body misalignments, and spatial artifacts, providing a purpose-built testbed for assessing robustness to generative errors. Extensive experiments demonstrate that our method achieves significant improvements in structural fidelity and temporal stability without any model training.
\end{abstract}
    
\section{Introduction}
\label{sec:intro}

Recent advances in large video diffusion transformers~\cite{yang2024cogvideox,ma2024latte,liu2024sora,peebles2023scalable} have brought unprecedented generative quality to text-to-video synthesis. By modeling joint spatial–temporal correlations through attention across both dimensions, these models can produce videos with coherent motion~\cite{yesiltepe2024motionshop, meral2026motionflow}, fine textures~\cite{dalva2024gantastic, dalva2025lorashop}, and diverse content~\cite{yesiltepe2025infinity, zheng2024stylebreeder,yesiltepe2024mist, yesiltepe2025dynamic} guided purely by textual input. Their scalability and learned priors make them powerful engines for video generation and editing. Yet, when such generative priors are applied to the task of restoration or super-resolution, they often expose a critical limitation: the tendency to replicate distorted evidence from the input rather than reconstructing underlying structure. In degraded or AI-generated footage, this behavior manifests as warped geometry, stretched motion, and temporally inconsistent details.

Conventional video super-resolution (VSR) methods~\cite{chan2021basicvsr,chan2022basicvsr++,liang2024vrt,geng2022rstt} rely on synthetic degradation pipelines and deterministic reconstruction losses. While effective on controlled benchmarks, they fail to generalize to real-world or generative content where degradations are complex and non-stationary. These models often sharpen artifacts rather than removing them, producing results that look crisp frame-wise but inconsistent across time. Diffusion-based VSR frameworks~\cite{he2024venhancer,zhou2024upscaleavideo,zhang2023i2vgen,yang2023mgldvsr} introduce powerful generative priors, yet the standard denoising formulation couples conditional and unconditional branches at the same timestep, forcing the model to remain overly faithful to corrupted low-resolution inputs. This coupling limits the ability of diffusion models to hallucinate plausible structure while maintaining temporal stability.

Several recent works have sought to overcome these limitations by refining temporal modeling or expanding the generative backbone. Upscale-A-Video~\cite{zhou2024upscaleavideo} enforces local and global temporal consistency through diffusion-based latent propagation, while VEnhancer~\cite{he2024venhancer} unifies spatial and temporal upsampling within a single diffusion transformer. STAR~\cite{xie2025star} incorporates local enhancement and frequency-aware losses to improve fidelity in real-world settings, and SeedVR2~\cite{wang2025seedvr2} accelerates restoration to a single sampling step using adversarial post-training. Despite their strong performance, these methods require extensive fine-tuning of large diffusion backbones and remain tied to specific training configurations, making them less flexible for arbitrary degradations or unseen content.

In this work, we present a training-free and model-agnostic framework for generative video super-resolution that operates entirely at inference time. Our key idea is to decouple the temporal evaluation of conditional and unconditional guidance within the diffusion sampling process. By querying the unconditional branch at a cleaner timestep closer to the data manifold and the conditional branch at the current timestep, the model receives a lookahead prior that discourages replication of spurious geometry while remaining anchored to the observed content. This simple temporal offset, which we term \emph{Decoupled Time Guidance} (DTG), progressively transitions the denoising process from structure correction to detail refinement, implicitly increasing the effective signal-to-noise ratio and stabilizing both geometry and appearance across frames.

After this geometry-preserving refinement, any off-the-shelf restoration or super-resolution module can be attached to further enhance high-frequency details. This plug-and-play composition preserves the structural corrections introduced by DTG while allowing specialized networks to focus solely on texture recovery. Together, these components form a practical pipeline that unifies generative reasoning with restoration fidelity, without retraining or architectural modification.

Our contributions are summarized as follows:
\begin{itemize}
    \item We propose \textbf{Decoupled Time Guidance} (DTG), a training-free modification to the diffusion sampling process that decouples conditional and unconditional denoising in time to preserve geometric integrity.
    \item We provide a theoretical interpretation showing that DTG implicitly adjusts the effective signal-to-noise ratio, yielding a cleaner, more stable diffusion trajectory.
    \item We demonstrate that DTG serves as a \textbf{plug-and-play} module applicable to any pretrained video diffusion transformer and can be followed by arbitrary restoration or super-resolution networks for high-frequency enhancement.
    \item Extensive experiments show that DTG improves structural fidelity, temporal consistency, and perceptual realism over recent diffusion-based video restoration approaches~\cite{he2024venhancer,zhou2024upscaleavideo,wang2025seedvr2,xie2025star}, while remaining computationally lightweight and fully model-agnostic.
\end{itemize}

\section{Related Work}
\label{sec:related}

\noindent \textbf{Text-to-Video Generation.}  
Recent progress in text-to-video (T2V) diffusion has been propelled by large-scale datasets and advances in diffusion transformer architectures. Early approaches adapted pre-trained image diffusion models by inserting temporal layers and fine-tuning them jointly on video data~\cite{chen2024panda,nan2024openvid,wang2023modelscope,blattmann2023stable,zhang2025show,wu2023tune,kara2024rave}, often balancing between preserving spatial priors and learning temporal coherence. The emergence of Diffusion Transformers (DiT)~\cite{peebles2023scalable} has since marked a paradigm shift, offering improved scalability and generative quality over 3D U-Nets, and now serves as the backbone for most state-of-the-art video diffusion frameworks~\cite{yang2024cogvideox,ma2024latte,liu2024sora}.  

Despite these advances, direct generation of high-resolution videos remains computationally expensive and limited by the availability of large, high-resolution video-text datasets. Consequently, modern T2V systems often adopt a cascaded design~\cite{wang2025lavie,zhang2023i2vgen,guo2024make}, where a base model first synthesizes low-resolution videos, followed by a refinement or super-resolution module that enhances spatial fidelity via a noise-guided reconstruction process~\cite{meng2021sdedit}. While effective, this pipeline requires holistic clip-level processing, incurring high memory cost and slow inference, which restricts scalability and real-world deployment.

\noindent \textbf{Generative Video Super-Resolution.}  
Video super-resolution (VSR) aims to reconstruct high-resolution (HR) frames from low-resolution (LR) inputs, restoring fine spatial details and temporal coherence. Conventional CNN- and Transformer-based architectures~\cite{chan2021basicvsr,chan2022basicvsr++,wang2019edvr,isobe2020video,liu2022learning,shi2022rethinking,geng2022rstt} largely depend on synthetic degradations and lack strong generative priors, which often leads to over-smoothing and suboptimal realism. To overcome this, diffusion-based generative VSR models have gained traction by leveraging pre-trained image and video diffusion priors. For instance, Upscale-A-Video~\cite{zhou2024upscaleavideo} introduces a latent diffusion framework with local-global temporal consistency mechanisms, achieving superior realism and temporal stability across both real-world and AI-generated videos. Similarly, VEnhancer~\cite{he2024venhancer} unifies spatial and temporal upsampling within a single diffusion model, using a video ControlNet and space-time augmentation to simultaneously enhance resolution and reduce flickering. These diffusion-based methods mark a transition from reconstruction-based restoration to generative enhancement, where perceptual quality, fidelity, and temporal coherence are jointly optimized.

\noindent \textbf{Generative Video Restoration.}  
While traditional VSR models perform well on synthetic benchmarks~\cite{chan2021basicvsr,liang2022recurrent,li2023simple,youk2024fma}, they generalize poorly to real-world or AI-generated content due to mismatch between training degradations and complex natural artifacts. Efforts like RealViformer~\cite{zhang2024realviformer} and RealBasicVSR~\cite{chan2022investigating} aim to bridge this gap through more diverse degradation modeling and architectural adaptations, but still lack strong generative capability for texture synthesis. More recently, diffusion-driven restoration frameworks~\cite{zhou2024upscaleavideo,he2024venhancer,yang2023mgldvsr} have demonstrated the potential of generative priors in reconstructing both photorealistic and temporally stable videos. Complementary approaches such as STAR~\cite{xie2025star} extend this idea further by integrating powerful text-to-video diffusion models into real-world VSR, enriching local details through spatial-temporal augmentation and frequency-aware learning.

\section{Methodology}
\label{sec:methodology}
\begin{figure*}
    \centering
    \includegraphics[width=1\linewidth]{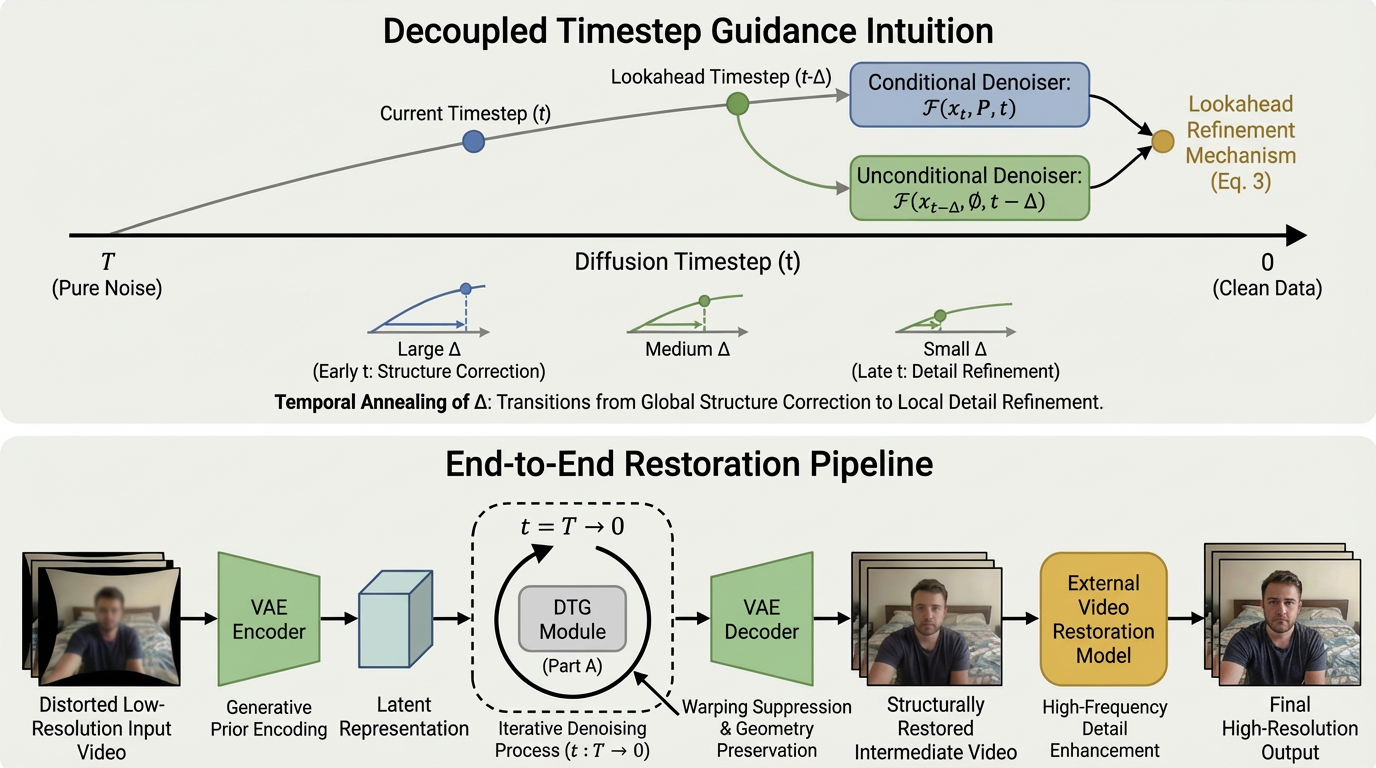}
    \vspace{-20pt}
    \caption{Overview of Decoupled Timestep Guidance (DTG). Given a distorted input video, DTG evaluates the unconditional denoiser at a cleaner timestep \(t-\Delta\) while keeping the conditional branch at the current timestep \(t\). This temporal decoupling provides a geometry-preserving lookahead prior that suppresses warped structures throughout sampling. After DTG refinement, any off-the-shelf video restoration model can be applied in a plug-and-play manner to recover high-frequency details, yielding stable and coherent outputs without retraining.
}
    \label{fig:framework}
    \vspace{-15pt}
\end{figure*}
\subsection{Preliminary: Video Diffusion Base Model}
\label{subsec:preliminary}

Our model is implemented on top of a pre-trained T2V diffusion model~\cite{wan2025wan}, which adopts the Rectified Flow framework~\cite{esser2024scaling} for the noise schedule and denoising process. The forward process is defined as straight paths between data distribution and a standard normal distribution, i.e.
\begin{equation}\label{eq:forward}
    z_t = (1-t)z_0 + t\epsilon,
\end{equation}
where $\epsilon \in \mathcal{N}(0,I)$ and $t$ denotes the iterative timestep.
To solve the denoising processing, we define a mapping between samples $z_1$ from a noise distribution $p_1$ to samples
$z_0$ from a data distribution $p_0$ in terms of an ordinary differential equation (ODE), namely:
\begin{equation}\label{eq:ODE}
dz_t=v_{\Theta}(z_t,t)dt, 
\end{equation}
where the velocity $v$ is parameterized by the weights $\Theta$ of a neural network. For inference, we employ Euler discretization for Eq.~\ref{eq:ODE} and perform discretization over the timestep interval at $[0, 1]$, starting at $t=1$. We then proceeed with iterative sampling with: $z_{t_i}=z_{t_j} + v_{\Theta}(z_{t_j},t_j) * (t_j-t_i)$.

For consideration of computing efficiency, the mapping is formulated in a latent space constructed with a 3D Variational Auto-Encoder (VAE)~\cite{kingma2013auto}. The core is a Transformer-based diffusion model (DiT)~\cite{peebles2023scalable}, where each block is instantiated as a sequence of spatial attention, temporal attention, and cross-attention modules. Text prompts and other micro parameters (like time step, aspect ratio, etc) are used as generative conditions. Due to the full-attention property of Transformer, our model only supports the generation of a fixed number of tokens as it was trained with, i.e. $512\times512$ of diverse aspect-ratio.
This is a common characteristic of almost all existing diffusion models for image or video generation~\cite{podell2023sdxl,chen2023videocrafter1,chen2024videocrafter2,yang2024cogvideox}, and extending pre-trained diffusion models for higher resolution generation remains an open research topic~\cite{he2023scalecrafter,guo2024make}. 

\subsection{Problem Definition}
\label{subsec:motivation}
Video restoration and super-resolution models that sharpen details often amplify warped faces, misalignments, stretched regions, and compression artifacts. These systems produce crisp frames but drift from perceptual fidelity and temporal consistency. We address this by separating structural and appearance signals in time: the unconditional branch is evaluated at a cleaner timestep near the data manifold, while the conditional branch remains at the current timestep. This lookahead prior suppresses spurious geometry while preserving fidelity, and an annealed temporal gap smoothly shifts guidance from structure recovery to detail refinement. The resulting schedule improves the effective signal-to-noise ratio, stabilizes content across frames, and hallucinates missing structure without retraining the backbone.

\subsection{Decoupled Time Guidance}
\label{subsec:lookahead_refinement}

\paragraph{Objective.}
Given a distorted low resolution conditioning sequence, our goal is to steer a pretrained video diffusion backbone toward geometrically plausible structure while preserving appearance. We achieve this by injecting a lookahead prior from a cleaner diffusion time into the current denoising step without any retraining.

\paragraph{Notation.}
Let \(F(x,t)\) denote the pretrained denoiser or flow field evaluated at diffusion time \(t\). At a current time index \(t\), define a cleaner anchor time
\[
\tau \coloneqq t - \Delta,\qquad 0 \le \Delta \le t,
\]
and an extrapolation coefficient \(\eta \in \mathbb{R}\). The forward process is variance preserving,
\[
x_t=\alpha_t x_0+\sigma_t \varepsilon,\qquad \mathrm{SNR}(t)=\alpha_t^{2} \big/ \sigma_t^{2}.
\]

\paragraph{Update rule.}
The Lookahead Refinement update performs an anchored extrapolation between the current and cleaner denoiser outputs
\begin{equation}
\label{eq:lr_update}
x^{\text{new}} \;=\; F(x,\tau) \;+\; \eta \bigl[F(x,t)-F(x,\tau)\bigr].
\end{equation}
For \(\eta=0\) the step snaps to the cleaner prediction \(F(x,\tau)\). For \(\eta=1\) it recovers the standard step at \(t\). For \(\eta>1\) it extrapolates beyond \(t\) along a direction anchored by the cleaner prior.

\paragraph{Score based interpretation.}
Using the Tweedie approximation \(F(x,t)\approx x+\sigma_t^{2}\nabla_x \log p_t(x)\) and substituting in \eqref{eq:lr_update} yields
\begin{align}
x^{\text{new}}
&\approx x + \Bigl[\sigma_\tau^{2} + \eta\bigl(\sigma_t^{2}-\sigma_\tau^{2}\bigr)\Bigr]\nabla_x \log p_t(x) \\
&\;\eqqcolon\; x + \sigma_{\text{eff}}^{2}\,\nabla_x \log p_t(x),
\end{align}
with the effective noise level
\begin{equation}
\label{eq:sigma_eff}
\sigma_{\text{eff}}^{2}
= \sigma_\tau^{2} + \eta\bigl(\sigma_t^{2}-\sigma_\tau^{2}\bigr).
\end{equation}
Since the signal component is anchored at \(\tau\), the corresponding effective signal to noise ratio is
\begin{equation}
\label{eq:snr_eff}
\mathrm{SNR}_{\text{eff}} = \frac{\alpha_\tau^{2}}{\sigma_{\text{eff}}^{2}}.
\end{equation}
Equations \eqref{eq:sigma_eff} and \eqref{eq:snr_eff} show that Lookahead Refinement behaves as denoising at an implicit time between \(\tau\) and \(t\), increasing the effective SNR when \(\eta>1\) and interpolating conservatively when \(0<\eta<1\).

\paragraph{Scheduling.}
We use a simple schedule
\begin{equation}
\Delta_t \searrow 0 \quad \text{as} \quad t \searrow 0,
\qquad
\eta_t \nearrow 1 \quad \text{as} \quad t \searrow 0,
\end{equation}
so that early steps draw structure from a cleaner anchor and later steps focus on detail refinement consistent with the observation.
\begin{figure*}[t]
    \centering
    \includegraphics[width=1\linewidth,trim=0 369 0 0,clip]{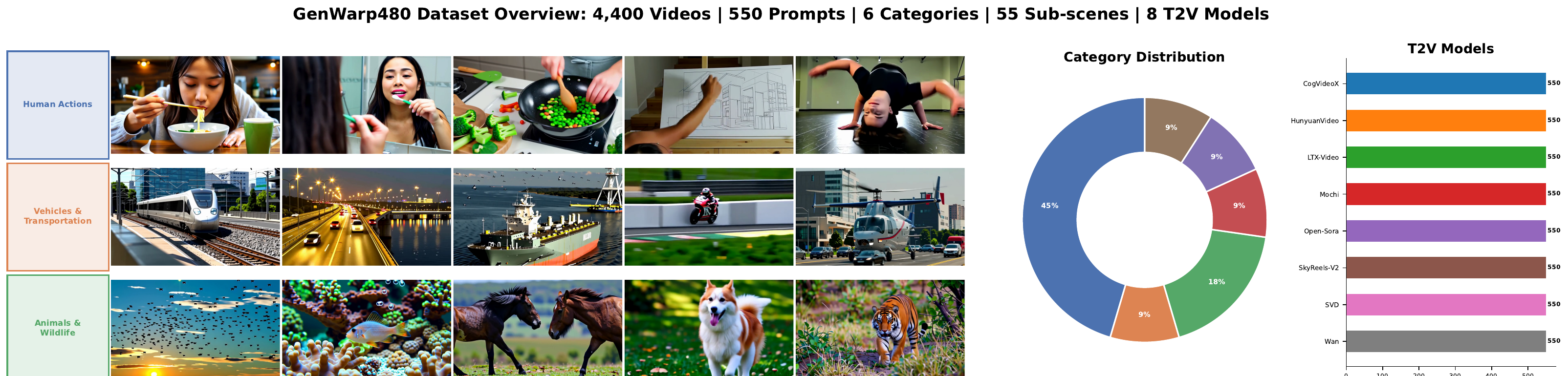}\\[-1mm]
    \includegraphics[width=1\linewidth,trim=0 0 0 369,clip]{figs/genwarp480.pdf}
    \vspace{-20pt}
    \caption{Sample frames and category distribution of the GenWarp480 dataset, covering diverse scenes and motions designed to evaluate temporal robustness and geometric stability in video upsampling.}

    \label{fig:genwarp}
    \vspace{-15pt}
\end{figure*}
\paragraph{Relation to classifier free guidance.}
Standard classifier free guidance combines conditional and unconditional predictions at the same time \(t\). Lookahead Refinement decouples these signals in time by anchoring one evaluation at \(\tau=t-\Delta\), which supplies a geometry preserving prior from a higher SNR state while the current time direction refines details.

\subsection{Plug and Play Detail Enhancement}
\label{subsec:plug_play_after_dtg}
After Decoupled Time Guidance corrects geometry and suppresses distortions, we attach any off the shelf restoration or super resolution model without retraining. Let \(\mathcal{T}_\phi\) denote Decoupled Time Guidance (DTG) and \(\mathcal{R}_\theta\) an arbitrary restoration module. The pipeline is
\[
\hat{x} = \mathcal{R}_\theta\!\left(x^{\text{new}}\,;\,y_{1:T}\right),
\]
where \(\mathcal{R}_\theta\) may optionally use external conditions \(y_{1:T}\) as side information. This composition is drop in, training free, and model agnostic. It preserves the structural corrections produced by DTG while allowing \(\mathcal{R}_\theta\) to focus on recovering high frequency detail.

\section{GenWarp480 Benchmark}
To enable systematic evaluation under realistic generative degradations, we introduce \textbf{GenWarp480}, a curated benchmark composed of AI-generated videos exhibiting diverse artifacts typical of modern text-to-video models. All videos are generated at a resolution of $480p$ with varied durations ranging from 3 to 5 seconds, rendered at 16 frames per second. The dataset encompasses a total of 4,400 clips distributed across multiple semantic categories to ensure both content diversity and temporal variety. Below, we describe the category design and content coverage in detail.

\subsection{Category Overview}
GenWarp480 is organized into six major categories, each designed to represent a unique visual and motion context commonly found in generative video synthesis. Each category contains multiple sub-scenes and prompt variations, ensuring a broad distribution of appearance, motion, and compositional complexity.

\noindent \textbf{1. Human Actions.} This category contains daily routine activities featuring clearly visible human subjects and natural face–body motion. Scenarios include eating at a table, drinking coffee or tea, brushing teeth in front of a mirror, folding laundry, cleaning, cooking, and performing simple exercises such as yoga or stretching. These clips challenge restoration models with fine-grained facial structures, hand–object interactions, and complex temporal dependencies.
\begin{figure*}[t]
    \centering
    \includegraphics[width=1\linewidth]{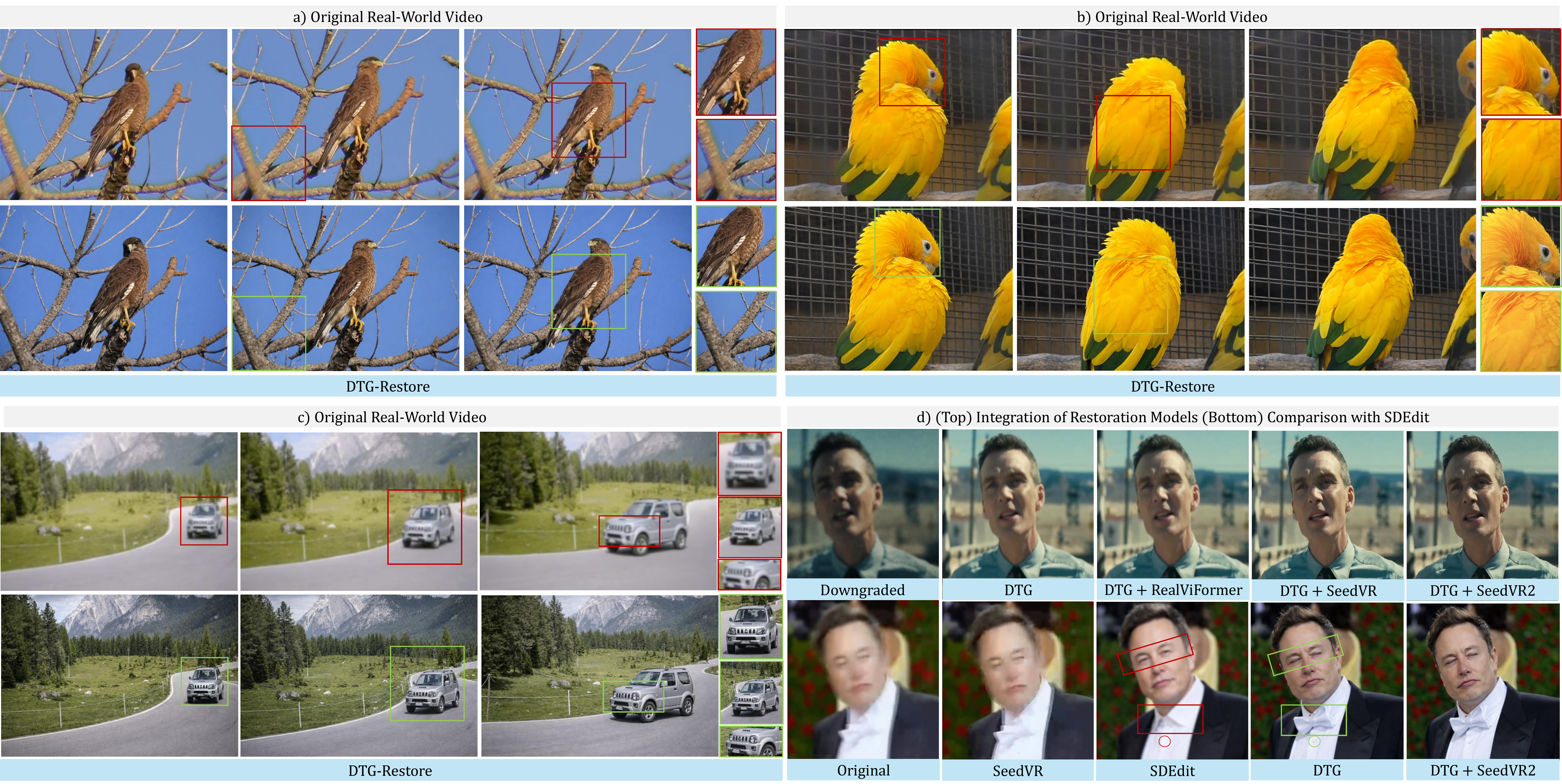}
    \vspace{-18pt}
    \caption{Additional qualitative comparisons on diverse scenes. Our method consistently improves geometric stability and visual coherence while suppressing distortion artifacts.}
    \label{fig:additional_results}
    \vspace{-12pt}
\end{figure*}
\begin{figure}[t]
    \centering
    \includegraphics[width=1\linewidth]{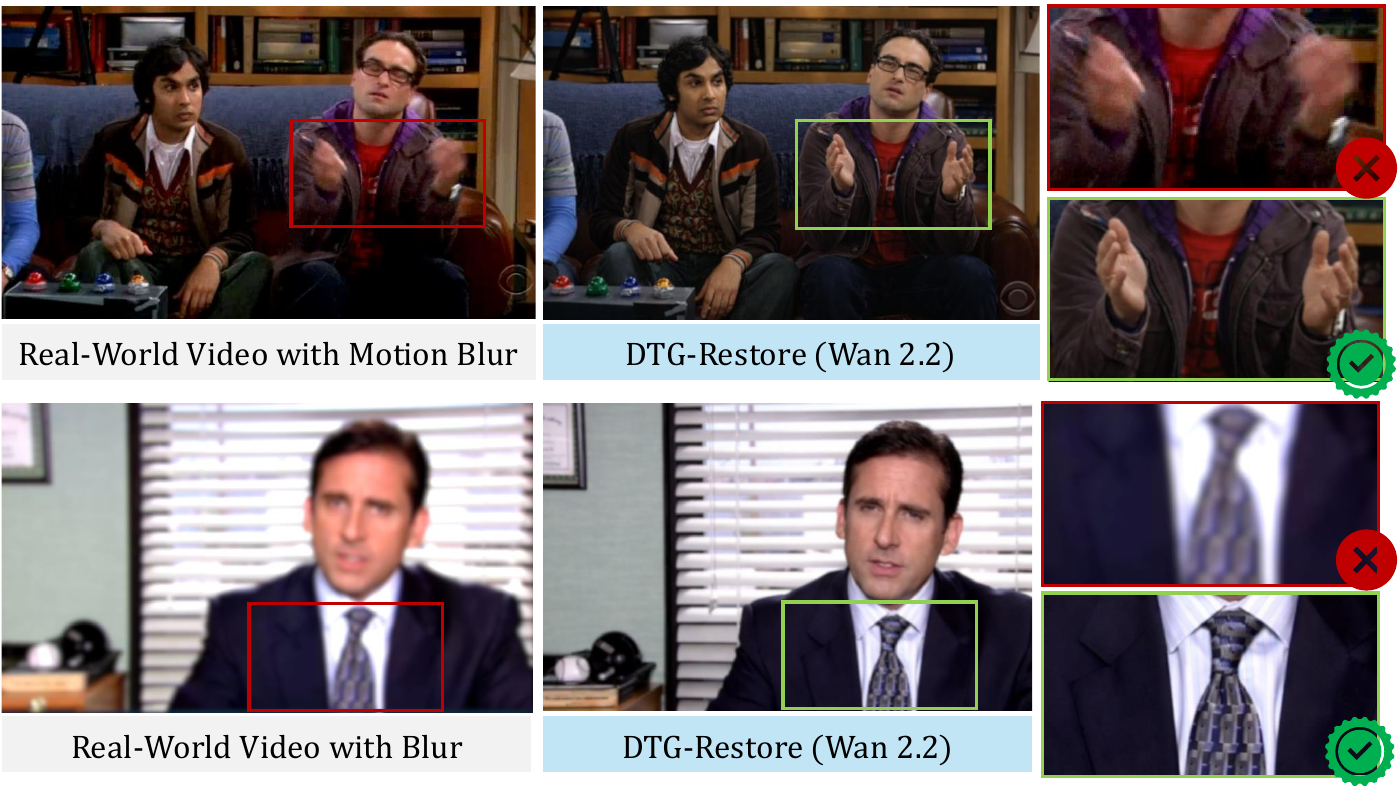}
    \vspace{-18pt}
    \caption{Real-world distortion cases. DTG-based enhancement restores cleaner details and more stable structures under challenging degradations.}
    \label{fig:real_world_results}
    \vspace{-10pt}
\end{figure}

\noindent \textbf{2. Nature and Environment.} This group includes dynamic natural scenes such as rivers flowing between rocks, waterfalls cascading down cliffs, ocean waves breaking on the shore, drifting clouds, volcanic eruptions, and falling snow in mountainous terrain. These videos exhibit fluid motion, particle dynamics, and light variations that are difficult for generative models to maintain consistently across frames.

\noindent \textbf{3. Animals and Wildlife.} This set features both terrestrial and aquatic animal motions, such as birds flying across the sky, fish swimming through coral reefs, dogs running in a park, cats climbing furniture, horses galloping in fields, and deer walking through forests. The fine motion details and organic textures in this category test a model’s capacity for preserving spatial coherence and biological motion realism.

\noindent \textbf{4. Vehicles and Transportation.} This category captures mechanical and man-made motion scenarios including airplanes taking off, trains moving through tunnels, cars driving in city traffic, ships sailing across the ocean, helicopters hovering over skylines, and motorbikes racing on tracks. These clips include high-speed motion, reflective surfaces, and frequent perspective changes—common sources of structural distortions in generative outputs.

\noindent \textbf{5. Urban and Architecture.} Urban scenes include cityscapes, night streets with neon lights, construction sites with cranes, bridges and highways from aerial views, and subway stations with moving trains. Architectural scenes often display flickering lighting, deformed geometry, and repetitive textures that expose weaknesses in temporal modeling. These videos serve as challenging cases for assessing geometric consistency and temporal alignment under structured spatial layouts.

\noindent \textbf{6. Objects and Daily Items.}
This category focuses on inanimate object motion and everyday indoor environments. Examples include clocks ticking on walls, candles flickering, balls bouncing, drones flying indoors, and robotic arms moving objects on a desk. Such scenes emphasize repetitive local motion and specular reflections, useful for evaluating fine-grained stability and texture reconstruction.

\definecolor{tabheader}{RGB}{246,248,252}
\definecolor{oursmetricbg}{RGB}{233,243,255}
\definecolor{oursrowbg}{RGB}{237,246,255}

\begin{table*}[t]
    \centering
    \caption{Quantitative comparisons on VSR benchmarks from diverse sources.}
    \vspace{-2mm}
    \label{tab:comparison}
    \small
    \setlength{\tabcolsep}{2.3mm}
    \renewcommand{\arraystretch}{1.15}
    \resizebox{\linewidth}{!}{
    \begin{tabular}{llccccccc>{\columncolor{oursmetricbg}}c}
    \toprule
    \rowcolor{tabheader}
    \multicolumn{2}{c}{\textbf{Dataset / Metric}} &
    \makecell[c]{RealViformer \\ \cite{zhang2024realviformer}} &
    \makecell[c]{MGLD-VSR \\ \cite{yang2023mgldvsr}} &
    \makecell[c]{UAV \\ \cite{zhou2024upscaleavideo}} &
    \makecell[c]{VEnhancer \\ \cite{he2024venhancer}} &
    \makecell[c]{STAR \\ \cite{xie2025star}} &
    \makecell[c]{SeedVR-7B \\ \cite{wang2025seedvr}} &
    \makecell[c]{SeedVR2-7B \\ \cite{wang2025seedvr2}} &
    \makecell[c]{\textbf{Ours}} \\
    \midrule
    \multirow{4}{*}{SPMCS}
        & PSNR $\uparrow$
        & \textbf{24.18} & \underline{23.39} & 21.68 & 18.52 & 22.59 & 20.73 & 20.66 & 22.76 \\
        & SSIM $\uparrow$
        & \textbf{0.658} & \underline{0.629} & 0.523 & 0.514 & 0.609 & 0.595 & 0.603 & 0.613 \\
        & LPIPS $\downarrow$
        & \textbf{0.382} & \underline{0.372} & 0.511 & 0.457 & 0.419 & 0.396 & 0.388 & 0.408 \\
        & DISTS $\downarrow$
        & 0.189 & \textbf{0.165} & 0.229 & 0.196 & 0.226 & 0.171 & \textbf{0.165} & \underline{0.169} \\
    \midrule
    \multirow{4}{*}{UDM10}
        & PSNR $\uparrow$
        & \textbf{26.78} & \underline{26.09} & 24.53 & 21.57 & 24.69 & 24.31 & 25.74 & 25.61 \\
        & SSIM $\uparrow$
        & \textbf{0.793} & 0.775 & 0.716 & 0.687 & 0.741 & 0.736 & \underline{0.783} & 0.703 \\
        & LPIPS $\downarrow$
        & 0.281 & 0.273 & 0.329 & 0.342 & 0.307 & \underline{0.267} & \textbf{0.225} & 0.271 \\
        & DISTS $\downarrow$
        & 0.165 & 0.147 & 0.185 & 0.177 & 0.189 & \underline{0.134} & \textbf{0.113} & 0.158 \\
    \midrule
    \multirow{4}{*}{REDS30}
        & PSNR $\uparrow$
        & \textbf{23.36} & 22.73 & 21.42 & 19.91 & 22.14 & 21.85 & 22.20 & \underline{23.12} \\
        & SSIM $\uparrow$
        & \textbf{0.617} & 0.579 & 0.524 & 0.551 & 0.599 & 0.589 & \underline{0.607} & 0.583 \\
        & LPIPS $\downarrow$
        & \underline{0.322} & \textbf{0.275} & 0.401 & 0.501 & 0.476 & 0.342 & 0.338 & 0.459 \\
        & DISTS $\downarrow$
        & 0.157 & \textbf{0.101} & 0.182 & 0.213 & 0.228 & 0.129 & \underline{0.126} & 0.136 \\
    \bottomrule
    \end{tabular}
    }
\end{table*}

\begin{table*}[t]
\centering
\small
\setlength{\tabcolsep}{10pt}
\renewcommand{\arraystretch}{1.2}
\caption{Quantitative comparison of perceptual metrics on the proposed benchmark. Metrics include LAION AP, MUSIQ, MANIQA, NIQE, and CLIP-IQA.}
\begin{tabular}{lccccc}
\toprule
\rowcolor{tabheader}
\textbf{Method} &
\textbf{LAION AP} $\uparrow$ &
\textbf{MUSIQ} $\uparrow$ &
\textbf{MANIQA} $\uparrow$ &
\textbf{NIQE} $\downarrow$ &
\textbf{CLIP-IQA} $\uparrow$ \\
\midrule
RealViformer       & 3.998 & \textbf{50.47} & 0.293 & \underline{4.014} & 0.482 \\
SeedVR             & 4.120 & 46.85         & \underline{0.278} & 4.128 & 0.496  \\
SeedVR2            & 4.423 & 37.28         & 0.242 & \textbf{3.915} & \underline{0.527} \\
VEnhancer          & 4.218 & 44.12         & 0.267 & 4.206 & 0.508  \\
Upscale-A-Video    & 4.371 & 45.67         & 0.273 & 4.198 & 0.517  \\
STAR               & \underline{4.457} & 41.96 & 0.261 & 4.263 & 0.418  \\
\rowcolor{oursrowbg}
Ours               & \textbf{4.642} & \underline{48.83} & \textbf{0.314} & 4.337 & \textbf{0.541} \\
\bottomrule
\end{tabular}
\label{tab:perceptual}
\end{table*}

 \section{Experiments}
\label{sec:exp}
\subsection{Evaluation Protocol}
 \noindent \textbf{Standard Benchmarks.}
We first assess our model's performance on established VSR datasets, including SMPCS, UDM10, RES30. These benchmarks are primarily designed to evaluate a model's ability to restore high-frequency details and upscale videos from low-resolution or blurry source inputs.

\textbf{Evaluation Metrics.} Our evaluation employs a dual-metric strategy tailored to the specific challenges posed by each type of dataset. For the standard VSR benchmarks, where a ground-truth high-resolution video is available, we measure restoration quality using established full-reference metrics. These include Peak Signal-to-Noise Ratio (\textit{PSNR}), Structural Similarity Index Measure (\textit{SSIM}), Learned Perceptual Image Patch Similarity (\textit{LPIPS}), and Deep Image Structure and Texture Similarity (\textit{DISTS}). These metrics collectively assess the pixel-level accuracy, structural fidelity, and perceptual similarity of the super-resolved videos against their ground-truth counterparts. Evaluating the enhancement of algorithmically generated videos requires a different approach, as the goal is to mitigate artifacts and improve perceptual quality rather than reconstructing a specific ground truth. For our curated dataset, we therefore focus on no-reference metrics that assess perceptual and aesthetic quality. We utilize a suite of metrics including LAION Aesthetic Predictor, MUSIQ, MANIQA, NIQE, and CLIP-IQA to quantify improvements in visual appeal and the reduction of distortions.

\noindent \textbf{Compared Methods.} We benchmark our method against a suite of recent and state-of-the-art video super-resolution models to demonstrate its competitive performance. The comparison includes the following methods: RealViformer, SeedVR, SeedVR-2B, VEnhancer, Upscale-A-Video, STAR, and MGLD-VSR. These models encompass a diverse range of architectural approaches and represent the forefront of video super-resolution research, thereby providing a rigorous baseline for our quantitative and qualitative comparisons. To ensure fair evaluation, all baselines are tested using their official implementations and recommended inference settings. We additionally include both GAN-based and diffusion-based pipelines to capture the full spectrum of reconstruction–generation trade-offs. This diversity allows us to assess robustness under varying motion, texture complexity, and degradation severities. Together, these benchmarks highlight the strengths and limitations of existing paradigms and contextualize the improvements introduced by our approach.

\begin{figure}[t]
    \centering
    \includegraphics[width=1\linewidth]{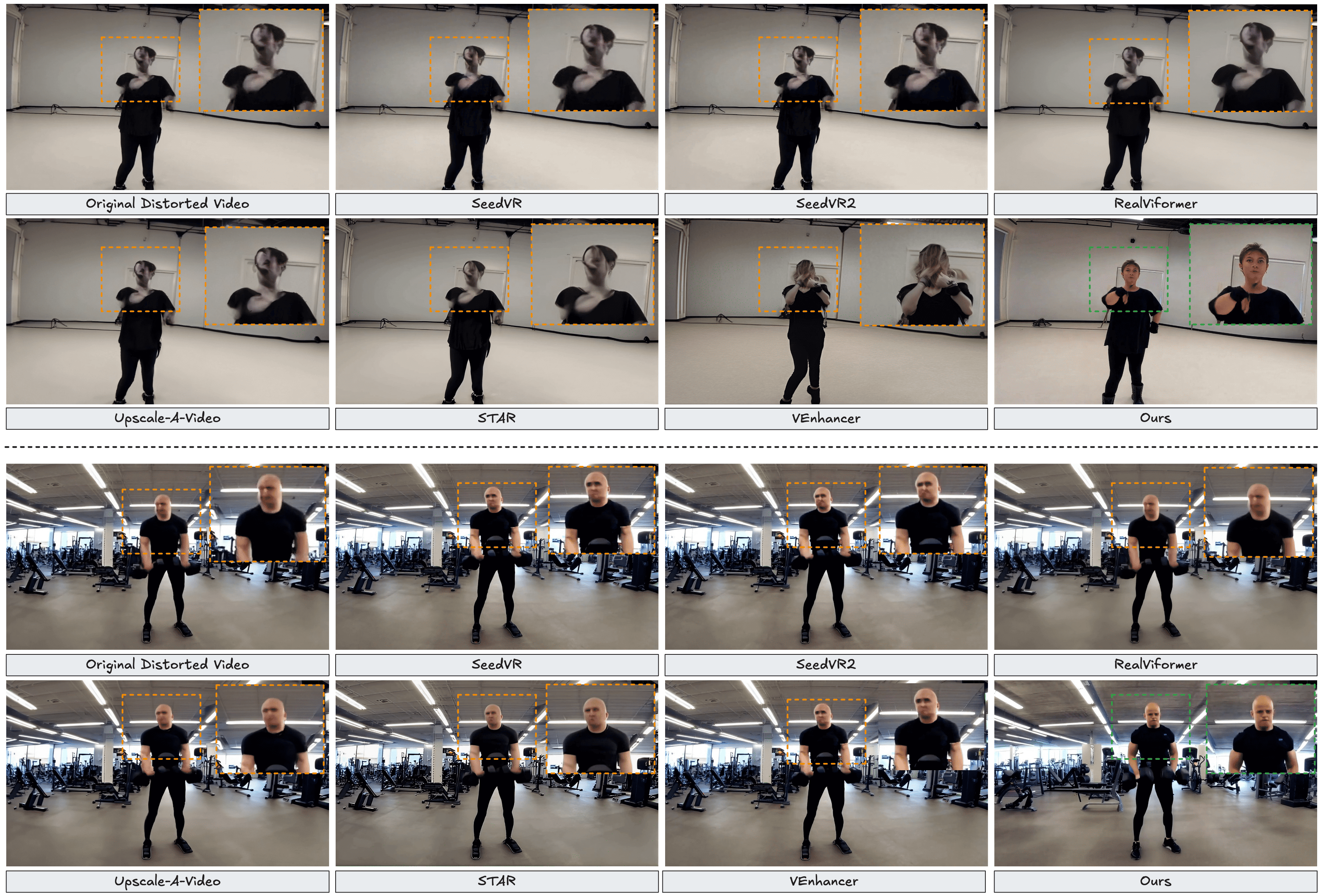}
    \vspace{-20pt}
    \caption{
Qualitative comparisons on distorted generative videos. Competing methods struggle with warped geometry and inconsistent motion, while our approach restores coherent structure and stable appearance.}
    \label{fig:comparison}
\end{figure}

\subsection{Main Results}
\textbf{Qualitative Results.} Figure~\ref{fig:comparison} compares our method with recent video enhancement and upsampling approaches on challenging real-world distorted inputs. Existing methods often amplify the spatial distortions already present in the low-quality sequence. SeedVR and SeedVR2 tend to magnify compression-related edge wobbling and background warping. RealViformer improves structural stability but still propagates the low-level artifacts visible in the original frames, particularly around limbs and facial regions. Upscale-A-Video and VEnhancer offer stronger generative priors yet frequently hallucinate inconsistent fine-scale textures that drift over time. STAR produces sharper details, but temporal windows occasionally introduce small discontinuities that degrade global coherence. In contrast, our approach preserves the underlying scene geometry without reinforcing the distortions. The proposed DTG-based creative upscaling reconstructs plausible high-frequency details while avoiding the propagation of visible artifacts. As shown in Figure~\ref{fig:comparison}, our method generates temporally stable results with cleaner boundaries and significantly fewer distortions around the hands, torso, and background structures. Additional qualitative examples are shown in Figures~\ref{fig:additional_results} and~\ref{fig:real_world_results}.

\noindent \textbf{Quantitative Results.}
We report quantitative comparisons on conventional restoration benchmarks in Table~\ref{tab:comparison}. Since these datasets evaluate pixel-level fidelity against a single ground truth, methods that enforce strict reconstruction objectives such as RealViformer~\cite{zhang2024realviformer} and MGLD-VSR~\cite{yang2023mgldvsr} achieve the highest PSNR, SSIM, LPIPS, and DISTS scores. In contrast, our framework is designed to prioritize creative consistency and perceptual realism through DTG rather than precise inversion of low-quality frames. As a result, our method does not aim to surpass reconstruction-centric models on PSNR-oriented datasets, and the scores in Table~\ref{tab:comparison} reflect this intended design choice. Nevertheless, our results show that we remain competitive across all metrics while producing reconstructions that are temporally stable and free from the distortion amplification observed in methods such as VEnhancer~\cite{he2024venhancer} and UAV~\cite{zhou2024upscaleavideo}. In particular, on SPMCS our method achieves a strong SSIM of 0.613, and on UDM10 we reach a perceptually favorable LPIPS of 0.271 despite not optimizing for pixel fidelity.

To more accurately assess the benefits of our method, we additionally evaluate all approaches on a curated reference-free benchmark tailored to creative and perceptual enhancement (Table~\ref{tab:perceptual}). These metrics measure realism, aesthetic quality rather than reconstruction similarity. Across this benchmark our method achieves the best overall performance, outperforming all baselines on LAION Aesthetic Predictor (4.642), CLIP-IQA (0.541), and MANIQA (0.314), while also obtaining the second-best MUSIQ score (48.83). SeedVR2~\cite{wang2025seedvr2} excels in NIQE due to its heavy smoothing, but performs poorly on other perceptual measures. STAR~\cite{xie2025star} performs competitively on LAION AP but falls behind significantly on CLIP-IQA.

These results confirm that while reconstruction-oriented metrics favor models with strong pixel-consistency bias, reference-free perceptual metrics clearly highlight the advantages of our creative upsampling strategy. Our method delivers higher visual realism, subject consistency, and artifact suppression.

\begin{table}[t]
\centering
\caption{\textbf{Ablation on $\Delta$ scheduling.} Eval. on GenWarp480 across diff. distortions measured by LAION AP and Sharpness score.}
\label{tab:delta_ablation}
\vspace{-5pt}
\resizebox{\columnwidth}{!}{%
\begin{tabular}{@{}lc|cc|cc|cc@{}}
\toprule
\multirow{2}{*}{\textbf{Schedule}} & \multirow{2}{*}{$\Delta_\text{end}$} & \multicolumn{2}{c|}{\textbf{Light}} & \multicolumn{2}{c|}{\textbf{Medium}} & \multicolumn{2}{c}{\textbf{Heavy}} \\
& & Quality$\uparrow$ & Sharp$\uparrow$ & Quality$\uparrow$ & Sharp$\uparrow$ & Quality$\uparrow$ & Sharp$\uparrow$ \\
\midrule
$\Delta{=}0$ (Std. CFG) & 0 & 4.12 & 0.768 & 4.15 & 0.761 & 3.91 & 0.739 \\
$\Delta{=}0.2$ (Const.) & 0.2 & 4.08 & 0.751 & 4.11 & 0.743 & 3.87 & 0.721 \\
$\Delta{=}0.3$ (Const.) & 0.3 & 4.03 & 0.724 & 4.06 & 0.716 & 3.82 & 0.694 \\
\midrule
Linear anneal & 0 & 4.51 & 0.812 & 4.48 & 0.804 & 4.41 & 0.791 \\
Cosine anneal & 0 & 4.56 & 0.824 & 4.53 & 0.817 & 4.47 & 0.806 \\
\rowcolor{blue!8}
\textbf{Ours (Exp.)} & 0 & \textbf{4.64} & \textbf{0.839} & \textbf{4.61} & \textbf{0.833} & \textbf{4.58} & \textbf{0.821} \\
\bottomrule
\end{tabular}%
}
\raggedright
\scriptsize
\vspace{-15pt}
\end{table}
\noindent \textbf{Ablation Studies.} Table~\ref{tab:SDEdit_ablation} compares DTG against SDEdit with different single-point resampling depths. DTG achieves the best results on all perceptual and geometric metrics: LAION AP (4.64), MANIQA (0.314), and CLIP-IQA (0.541), while also obtaining the lowest warp score (0.071). The strongest SDEdit setting ($t_{\text{start}}{=}0.5$) reaches LAION AP 4.38 and CLIP-IQA 0.512, but still trails DTG by a clear margin and yields worse geometry (warp 0.118). We also observe the expected SDEdit trade-off: shallow resampling ($t_{\text{start}}{=}0.3$) preserves less structure (warp 0.142), while deeper resampling ($t_{\text{start}}{=}0.7$) reduces warp (0.097) but noticeably hurts perceptual quality (LAION AP 4.29, CLIP-IQA 0.478). In contrast, DTG's per-step decoupling improves both realism and structural stability simultaneously.

Table~\ref{tab:delta_ablation} further analyzes how the $\Delta$ schedule affects robustness under light, medium, and heavy distortions. Constant $\Delta$ schedules underperform and degrade as $\Delta$ increases, indicating that fixed lookahead is too rigid across timesteps. Annealed schedules are substantially better, with cosine annealing outperforming linear annealing in all three distortion regimes. Our exponential schedule performs best overall, achieving the highest quality and sharpness for each regime (e.g., under heavy distortion: 4.58 / 0.821 vs. 4.47 / 0.806 for cosine and 3.91 / 0.739 for standard CFG with $\Delta{=}0$). These results validate that a strong early lookahead followed by progressive decay is critical for balancing geometric correction and detail recovery.



\begin{table}[t]
\caption{\textbf{Comparison against SDEdit.} DTG's per-step decoupling surpasses SDEdit's single-point re-sampling across perceptual quality and geometric fidelity metrics.}
\label{tab:SDEdit_ablation}
\vspace{-5pt}
\centering
\small
\setlength{\tabcolsep}{3pt}
\begin{tabular}{lcccc}
\toprule
Method & LAION$\uparrow$ & MANIQA$\uparrow$ & CLIP-IQA$\uparrow$ & Warp$\downarrow$ \\
\midrule
SDEdit ($t_{\text{start}}$=0.3) & 4.21 & 0.267 & 0.489 & 0.142 \\
SDEdit ($t_{\text{start}}$=0.5) & 4.38 & 0.281 & 0.512 & 0.118 \\
SDEdit ($t_{\text{start}}$=0.7) & 4.29 & 0.258 & 0.478 & 0.097 \\
\rowcolor{blue!8}
\textbf{DTG (Ours)} & \textbf{4.64} & \textbf{0.314} & \textbf{0.541} & \textbf{0.071} \\
\bottomrule
\end{tabular}
\vspace{-12pt}
\end{table}

\begin{table}[h]
\centering
\scriptsize
\setlength{\tabcolsep}{4pt}
\resizebox{\columnwidth}{!}{%
\begin{tabular}{lccc}
\toprule
\shortstack{\textbf{Method}} &
\shortstack{\textbf{Sharpness}\\\textbf{Clearness}} &
\shortstack{\textbf{Motion}\\\textbf{Smoothness}} &
\shortstack{\textbf{Aesthetic}\\\textbf{Quality}} \\
\midrule
SeedVR2          & 3.80 & 3.32 & 3.60 \\
RealViformer     & 3.52 & 3.46 & 3.58 \\
STAR             & 2.92 & 2.90 & 3.10 \\
Upscale-A-Video  & 2.84 & 2.76 & 2.84 \\
Venhancer        & 3.84 & 3.80 & 3.88 \\
\rowcolor{blue!8}
DTG-Restore (Ours) & \textbf{4.40} & \textbf{4.52} & \textbf{4.36} \\
\bottomrule
\end{tabular}%
}
\caption{\textbf{User Study on Perceptual Quality.} Participants evaluate each method on Sharpness/Clearness, Motion/Smoothness, and Aesthetic/Quality through a preference experiment on generated videos.}
\label{tab:user_study}
\end{table}
\noindent{\textbf{User Study.}} A user study with 50 participants evaluated 60 generated videos using a standardized interface. Each video was rated on a 1–5 Likert scale for clarity/sharpness, motion smoothness, and overall aesthetic quality. Results in Table \ref{tab:user_study} show that DTG-Restore achieved the highest scores across all metrics, indicating consistent user preference. Venhancer ranked second but with a noticeable gap, especially in motion quality. SeedVR2 performed well in sharpness but lacked motion stability, while RealViformer showed balanced yet less competitive results. STAR and Upscale-A-Video scored lowest due to temporal artifacts and inconsistencies. Overall, DTG-Restore aligns strongly with human subjective preferences and outperforms competing methods.
\section{Limitations and Conclusion}
While our approach demonstrates strong creative upsampling capabilities, it remains inherently dependent on the representational biases and reconstruction limits of the pretrained diffusion backbone. Since our framework does not retrain the underlying generative model, failure cases may arise in regions where the base model lacks sufficient priors or where degradation is severe. Moreover, the hallucination–reconstruction balance can still be challenging in extreme motions or textures that deviate from the pretrained model’s distribution. Despite these constraints, our results show that structured creative upsampling within a diffusion–transformer architecture provides a robust path toward high-quality, temporally coherent video enhancement.

{
    \small
    \bibliographystyle{ieeenat_fullname}
    \bibliography{main}
}
\newpage
\appendix
\setcounter{page}{1}
\maketitlesupplementary

\section{Details on User Study}
\begin{figure}
    \centering
    \includegraphics[width=1\linewidth]{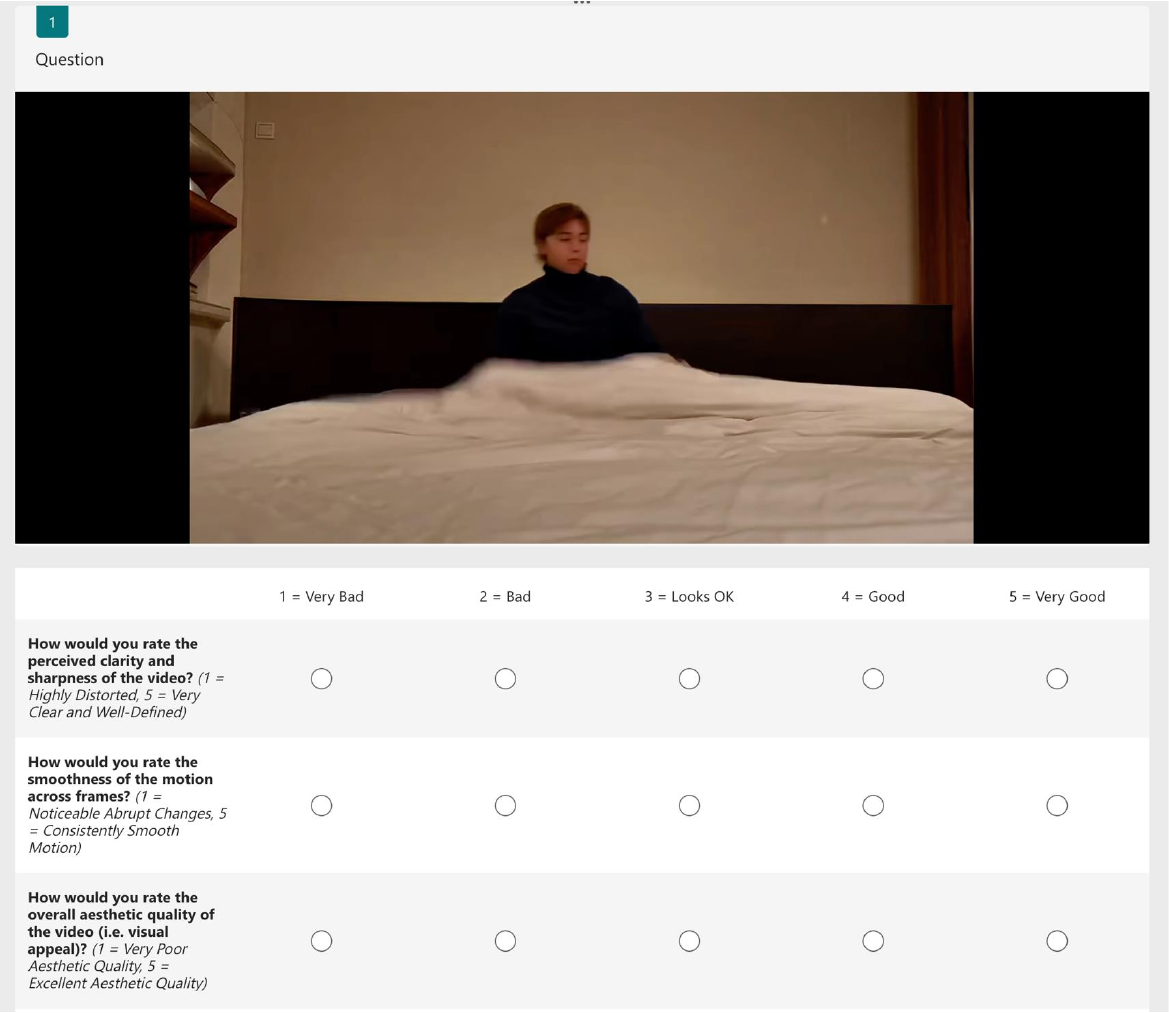}
    \caption{Example of the user study interface used in our evaluation. Each participant watched one video at a time and subsequently rated its sharpness/clearness, motion smoothness, and overall aesthetic quality using a 1--5 Likert scale.}

    \label{fig:user_study_creativity}
\end{figure}

We conducted a user study with 50 participants who evaluated a total of 60 generated videos using a standardized interface, an example of which is shown in Fig.~\ref{fig:user_study_creativity}. Participants viewed each video independently and answered three specific questions on a 1--5 Likert scale: \textbf{(i) ``How would you rate the perceived clarity and sharpness of the video?''} \textbf{(ii) ``How would you rate the smoothness of the motion across frames?''} \textbf{(iii) ``How would you rate the overall aesthetic quality of the video (i.e., visual appeal)?''} The scoring interface explicitly clarified the meaning of each score range to ensure consistent evaluations across users (see page~1 of the provided questionnaire form). The aggregated results are reported in Table~\ref{tab:user_study}, where our DTG-Restore achieves the highest scores across all three metrics, indicating that it was consistently preferred by participants. Venhancer follows as the second-best method but maintains a notable gap from ours, particularly in motion-related quality. SeedVR2 demonstrates strong sharpness but shows instability in motion smoothness, whereas RealViformer presents more balanced but less competitive results. STAR and Upscale-A-Video receive the lowest scores, reflecting their struggle with temporal artifacts and perceptual inconsistencies. These findings confirm that DTG-Restore not only achieves state-of-the-art quantitative results in the creative upscaling task but also aligns strongly with human subjective preferences.

\section{Additional Ablations}
\begin{figure*}
    \centering
    \includegraphics[width=1\linewidth]{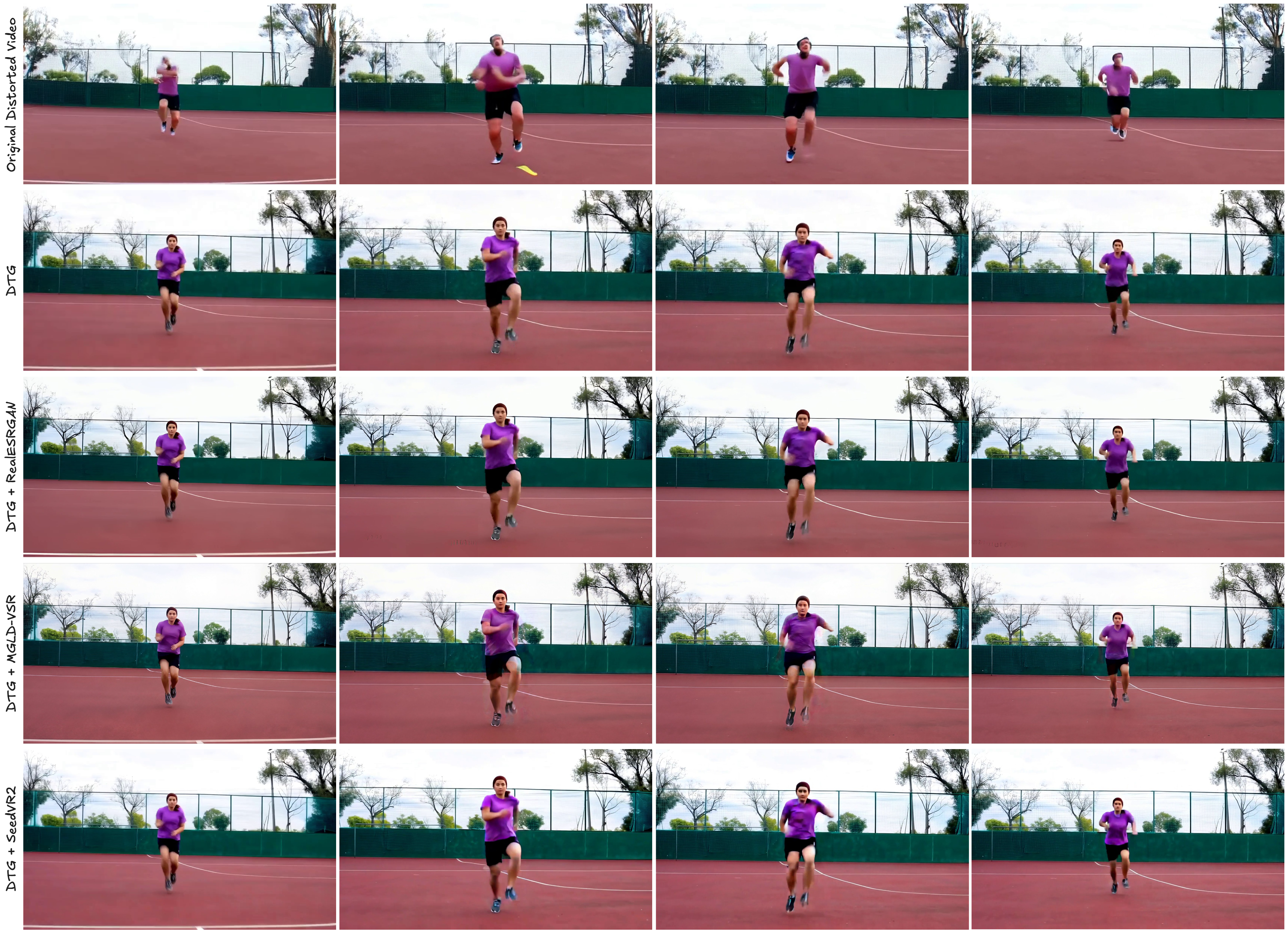}
    \vspace{-20pt}
    \caption{Ablation study of DTG and its plug-and-play detail modules. DTG alone corrects geometry and stabilizes motion, while combining DTG with various restoration models further enhances fine details.}

    \label{fig:ablation_plug_play}
    \vspace{-15pt}
\end{figure*}

Figure~\ref{fig:ablation_plug_play} presents an ablation on DTG and its plug-and-play detail enhancement stage. DTG alone already corrects large geometric distortions and improves temporal stability, indicating that the core gains come from decoupled time guidance during sampling. When DTG is followed by restoration backbones, fine textures and local details are further improved while preserving the corrected structure. This trend is consistent across diverse scenes, showing that DTG serves as a robust first-stage geometry prior and that downstream enhancers primarily contribute detail refinement rather than structural repair.

\end{document}